\pdfoutput=1

\documentclass[11pt]{article}

\usepackage[]{naacl2021}

\usepackage{times}
\usepackage{latexsym}

\usepackage[T1]{fontenc}

\usepackage[utf8]{inputenc}

\usepackage{microtype}

\usepackage{enumitem}
\setlist[itemize]{noitemsep, nolistsep}
\usepackage{amsfonts}
\usepackage{amsmath}
\usepackage{booktabs}
\usepackage{multirow}
\usepackage{graphicx}
\usepackage{caption}
\usepackage{subcaption}
\usepackage{hyperref}
\usepackage[symbol]{footmisc}
\usepackage[ruled,vlined]{algorithm2e}
\usepackage{algorithmic}

\title{\textit{RefineCap}: Concept-Aware Refinement for Image Captioning}

\author{Yekun Chai$^\dagger$, Shuo Jin$^\ddagger$,  Junliang Xing$^\dagger$\\
      $^\dagger$Institute of Automation, Chinese Academy of Sciences \\
      $^\ddagger$University of Pittsburgh\\
  \texttt{chaiyekun@gmail.com} \quad \texttt{shj42@pitt.edu} \quad \texttt{jlxing@nlpr.ia.ac.cn}\\
  }
  
\begin{document}
\maketitle
\begin{abstract}
Automatically translating images to texts involves image scene understanding and language modeling. In this paper, we propose a novel model, termed \emph{RefineCap}, that refines the output vocabulary of the language decoder using decoder-guided visual semantics, and implicitly learns the mapping between visual tag words and images. The proposed Visual-Concept Refinement method can allow the generator to attend to semantic details in the image, thereby generating more semantically descriptive captions. Our model achieves superior performance on the MS-COCO dataset in comparison with previous visual-concept based models.
\end{abstract}

\section{Introduction}
\label{sec:intro}
Holding the promise of bridging the domain gap between computer vision and human language, image-to-text translation, \emph{a.k.a} image captioning, has lately received great attention in both communities~\cite{johnson2016densecap,chen2017sca,anderson2018bottom,fan2019bridging}. Combining image scene understanding and language generation, it aims to translate descriptive texts given corresponding images.


%




Existing work maintained that frequently occurred n-grams of reference captions in the training set are preferred in the caption generation, regardless of image contents~\cite{fan2019bridging}. Visual concept (\emph{i.e., tag}) prediction is proposed to leverage the visual semantics from images for generating relevant words~\cite{wu2016value,you2016image,gan2017semantic,yao2017incorporating,fan2019bridging}. It predicts the probability of each semantic concept that occurs in the corresponding image out of the selected image-grounded vocabulary from reference captions, for the use of following caption generation. 

However, most of them adopted Long Short-Term Memory (LSTM)~\cite{hochreiter1997long} as the language decoder, in which sequence-align recurrence inherently precludes parallelization in practice. Initially proposed for neural machine translation (NMT), Transformer~\cite{vaswani2017attention} architecture has made rapid progress in numerous applications, such as NMT~\cite{ott2018scaling,gu2019levenshtein}, image generation~\cite{parmar2018image}, automatic speech recognition~\cite{dong2018speech}, computer games~\cite{vinyals2019grandmaster}, \emph{etc}. We adopt Transformer blocks rather than RNN to support parallel training in our model.


\begin{figure}[t]
\vskip 0mm
\begin{center}
\includegraphics[width=\columnwidth]{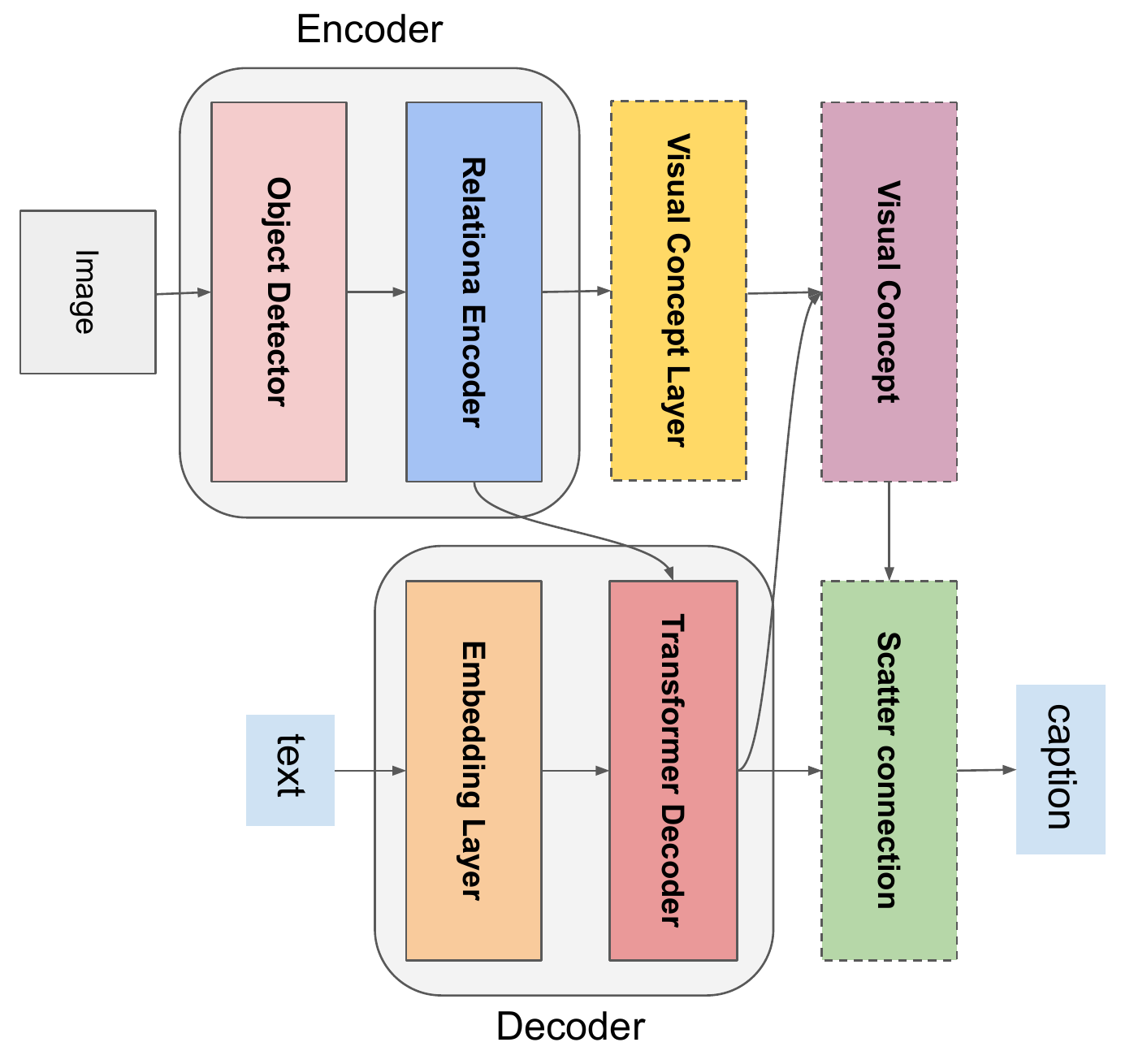}
\vskip -2mm
\caption{Schematic illustration of proposed models, where ``text'' indicates previous words of the captions.}
\label{fig:arch}
\end{center}
\vskip -8mm
\end{figure}

Intuitively, when describing an image, people often take much time to ``think'' of the words to generate coherent functional words, and just ``look'' at the image regions occasionally for content words. Similarly, for image captioning, only content-related words are actively matched with the image, whereas other functional words can often be automatically inferred using a language model.

The adaptive attention model~\cite{lu2017knowing} found that not all generated words are actively related to visual contents—some words can be reliably predicted just from the language model, such as ``phone'' after ``taking on a cell''. Also, non-visual words like ``the'' can be generated with the language model inference. 

Grounding on this, we introduce a language-decoder-guided gate to regulate visual-concept vectors, followed by a scatter connection layer to map each element of the image-grounded vector to the corresponding position in the decoder's vocabulary. This jointly takes into account what the captioner ``thinks'' via the language model and what it ``looks'' at with the dynamic visual-semantic concept vector. In this paper, we propose \emph{RefineCap}, a visual-concept-aware refined encoder-decoder architecture to dynamically modulate the visual-semantic representations and thus enhance the output of language model. Within the proposed model, visual signals are decoder-regulated and the content-based gate removes the independent assumption of visual objects in the attention mechanism.




Our work illustrates that image-grounded concept detection advances the performance of Transformer-based encoder-decoder captioning architecture by incorporating the visual-semantic representation using reinforcement learning. Our key contributions are as follows:

\setlist{nolistsep}
\begin{itemize} [noitemsep]
    \item A scatter-connected mechanism to refine the language decoder using extracted visual semantics, which produces more specific descriptive words in caption generation.
    \item A competing model that outperforms the previous visual-concept based captioning models on the MS-COCO dataset.
\end{itemize}

\section{Methodology}
\label{sec:method}
Fig.~\ref{fig:arch} shows an overview of the proposed model. We use an object detector to extract visual object features and a relational encoder to capture the pairwise relations among detected objects (Sec.~\ref{subsec:enc}). Then visual-concept components learn the implicit visual tags in the image. The Transformer language decoder receives the extracted object features (Sec.~\ref{subsec:dec}), followed by a scatter connection mechanism (Sec.~\ref{subsec:refine}) for decoder refinement.

\subsection{Transformer as Relational Encoder}
\label{subsec:enc}
Following~\citep{anderson2018bottom}, we use Faster RCNN~\cite{ren2015faster} with ResNet-101~\cite{he2016deep} as the object detector. We extract the object proposals using Region Proposal Network and mean-pooled convolution to generate the 2048-dimensional proposal feature. 

Let $\mathbf{X} = [ \mathbf{x}_1  \mathbf{x}_2 \cdots \mathbf{x}_M]^\top \in \mathbb{R}^{M \times 2048}$ denote the extracted $M$ visual proposal features of each image. We use a fully connected layer to reduce the spatial feature of 2048 into $D=512$. Then we apply the standard Transformer encoder with layer number $N_\textrm{enc}$ to capture the object-wise relation as in~\cite{zambaldi2018deep}. The output of encoder is represented as $\mathbf{F} = [ \mathbf{f}_1  \mathbf{f}_2 \cdots \mathbf{f}_M]^\top \in \mathbb{R}^{M \times D}$.

\subsection{Transformer Decoder}
\label{subsec:dec}
Given the caption sequence with the length of $T$, we apply a standard Transformer decoder of $N_\textrm{dec}$ layers as the caption generator. The input sequence firstly passes into word embedding and sinusoidal positional embedding layers, followed by a masked self-attention sublayer attending to previous histories. Then a cross attention sublayer as in ~\citet{vaswani2017attention} is applied to capture the multi-modal attention between each word and extracted object features, followed by a position-wise feed-forward layer (FFN). All sublayers are encompassed by a residual connection~\citep{he2016deep} and layer normalization~\citep{ba2016layer}.

\subsection{Visual-Concept Refinement}
\label{subsec:refine}
\paragraph{Visual Concept Layer} The visual concept layer is designed to extract the probability of common concept words, such as noun and verbs. We employ the $K$ most frequently occurring words whose POS tags are nouns, verbs, or adjectives as the image-grounded concept vocabulary set $\mathcal{V}_\text{tag}$. Following~\cite{gan2017semantic}, $K$ is set to 1,000 in our experiment. For each image, the visual concept layer projects its visual object features $\mathbf{f}_m$ ($m=1,\cdots,M$) into a $(K/M)$-dimensional vector, and then concatenates $M$ different visual outputs to get the $K$-dimensional concept representation. Then we apply an activation function to get the probability of each concept word in $\mathcal{V}_\text{tag}$, which is a multi-label binary prediction.
\begin{align}
    \mathbf{v} &{}= \mathbf{f}_1\mathbf{W_0} \Vert \mathbf{f}_2\mathbf{W_0} \Vert \cdots\Vert \mathbf{f}_M\mathbf{W_0} \\
    \hat{\mathbf{v}} &{}= \sigma (\mathbf{v}) \label{eq:tag_loss}
\end{align}{where $ \mathbf{W_0} \in \mathbb{R}^{D \times K/M }$ denotes trainable weights, $\Vert$ denotes the concatenation along the last axis, $\sigma (x) = 1 / (1 + \exp (-x))$, $\mathbf{f}_m \in \mathbb{R}^D$ ($m=1,2,\cdots,M$) is the $m$-th object feature out of $M$ visual proposals.  $\hat{\mathbf{v}} \in \mathbb{R}^K$ represents the visual concept vector, in which each element represents the confidence of image-grounded concept words.}

Denoting the decoder output at $t$-th time step ($t = 1,2, \cdots, T$) by $\mathbf{h^t} \in \mathbb{R}^{D}$, we compute the context vector $\mathbf{c^t}$ by considering the interaction between the $t$-th decoder output and all encoded features $\mathbf{F}$ in one image, followed by a non-linearity $g$ (we use sigmoid function here) to produce the decoder-guided gate for concept-aware modulation.
\begin{align}
    \mathbf{c^t} &{}= \mathbf{u}^\top \tanh(\mathbf{W_1 h^t} + \mathbf{W_2 \mathbf{F}}) \\
    \mathbf{g^t} &{} = g(\mathbf{W_3} \mathbf{c^t})
    \end{align}{where $ \{ \mathbf{W_1}, \mathbf{W_2}\} \in \mathbb{R}^{D^\prime \times D}, \mathbf{u} \in \mathbb{R}^{D^\prime}, \mathbf{W_3} \in \mathbb{R}^{K \times M}$ are parameters, $D^\prime$ represents the intermediate hidden dimension.}

At $t$-th time step, the visual-concept vector $\hat{\mathbf{v}}$ is modulated by the decoder-guided gate $\mathbf{g^t}$ to render the final refined representation $\mathbf{o^t}$:
\begin{align}
    \mathbf{o^t} = \mathbf{g^t} \odot \hat{\mathbf{v}}
\end{align}{where $\odot$ indicates the element-wise product.}

\paragraph{Scatter-Connected Mapping} Since the selected image-grounded vocabulary $\mathcal{V}_\text{tag}$ is the subset of caption vocabulary $\mathcal{V}_\text{cap}$, \emph{i.e.}, $\mathcal{V}_\text{tag} \subset \mathcal{V}_\text{cap}$, we apply scatter-connected mapping by adding the corresponding element of $\mathcal{V}_\text{tag}$ onto $\mathcal{V}_\text{cap}$ to enhance the confidence of concept word prediction:

\begin{align}
    \mathbf{h^t} [j] =\left\{
                \begin{array}{ll}
                  \mathbf{h^t} [j]  +  \mathbf{o^t} [k] & \textrm{if } \mathcal{V}_\text{cap}(j) = \mathcal{V}_\text{tag}(k)\\
                    \mathbf{h^t} [j] & \textrm{otherwise},
                \end{array}
              \right.
\end{align}{where $\mathcal{V}_\text{cap}(j)$ and $\mathcal{V}_\text{tag}(k)$ indicate the corresponding concept in the $j$-th position of caption vocabulary set and $k$-th word of concept vocabulary set. $[.]$ is tensor indexing operation.}
Then a softmax function is applied afterward to get the probability over all caption vocabularies.

\subsection{Training with Policy gradient}
\label{subsec:train}

\paragraph{Captioning as Reinforcement Learning} The image captioning task is cast to the RL problem: the policy network \emph{RefineCap}, defined as $\pi_\theta$ parameterized by $\theta$, takes an action $a_t$ at $t$-th time step for each observation (\emph{i.e.}, image) to predict the next word $w_t$ until reaching the rollout end. The return $G$ for each rollout is defined as CIDEr-D scores~\cite{vedantam2015cider} between hypothesis and ground-truth captions.

We leverage the REINFORCE with baseline algorithm to reduce the gradient variation. The parameter $\theta_{t+1}$ is updated as follows:
\begin{align}
    \theta_{t+1} &{}= \theta_t + \alpha (G_t - b) \nabla \ln \pi(\cdot)
\end{align}{where $b$ takes the average of batch returns, $\alpha$ is the learning rate.}



\section{Experiments}
\label{sec:exp}
\subsection{Experimental Setup}
\paragraph{Dataset and Evaluation} We experiment on the MS-COCO dataset~\cite{lin2014microsoft} and report the performance on \textit{Karpathy} offline splits, consisting of 113,287/5,000/5,000 images for training/val/test sets, in which each image is paired with 5 human annotations. We empploy BLEU~\cite{papineni2002bleu}, ROUGE-L~\cite{Lin2004ROUGEAP}, METEOR~\cite{denkowski2014meteor}, CIDEr-D~\cite{vedantam2015cider}, and SPICE~\cite{anderson2016spice} as evaluation metrics.

\paragraph{Implementation Details} 
We set the embedding dimension $D$ to 512, the layer numbers of both encoder and decoder as 3 (for fast training), batch size of 50, head number $h$ as 8, the hidden dimension of FFN layer as 2,048, the maximum number of extracted features as 50. Word embeddings are randomly initialized. We employ Adam optimizer~\cite{kingma2014adam} with $\beta_1=0.9, \beta_2=0.98$ as in~\cite{li2019entangled}. To avoid over-fitting, we set the dropout rate as 0.1, and early stopping patience as 5 during training. The beam width is set as 5 during beam search decoding. To initialize the model weights for RL training, we pretrain the model with supervised learning using cross-entropy loss with the same setting. All experiment are trained on a single NVIDIA Tesla V100 GPU.

\subsection{Results}
\label{subsec:res}
Table~\ref{tab:results} exhibits the performance of visual-concept based image-to-text translation models on MS-COCO dataset, in which the proposed model outperforms baseline models by a clear margin.

\begin{table*}[thb]
\vskip -5mm
\resizebox{\linewidth}{!}{%
\begin{tabular}{@{}lllllllll@{}}
\toprule
Models & BLEU-1 & BLEU-2 & BLEU-3 & BLEU-4 & METER & ROUGE-L & CIDEr & SPICE \\ \midrule
SemAttn~\cite{you2016image} & 0.709 & 0.537 & 0.402 & 0.304 & 0.243 & - & - & - \\
Att-CNN+LSTM~\cite{wu2016value} & 0.74 & 0.56 & 0.42 & 0.31 & 0.26 & - & 0.94 & - \\
LSTM-C~\cite{yao2017incorporating} & - & - & - & - & - & 0.230 & - & - \\
Skeleton Key~\cite{wang2017skeleton} & 0.673 & 0.489 & 0.355 & 0.259 & 0.247 & 0.489 & 0.966 & 0.196 \\
SCN-LSTM~\cite{gan2017semantic} & 0.728 & 0.566 & 0.433 & 0.330 & 0.257 & - & 1.041 & - \\
Bridging~\cite{fan2019bridging} & - & - & - & 0.330 & 0.264 & \textbf{0.586} & 1.066 & - \\
Ours & \textbf{0.802} & \textbf{0.645} & \textbf{0.499} & \textbf{0.378} & \textbf{0.283} & 0.580 & \textbf{1.272} & \textbf{0.225} \\ \bottomrule
\end{tabular}%
}
\caption{Overall performance of the proposed model and \textbf{visual-concept based} models.}
\label{tab:results}
\end{table*}

\paragraph{Ablation Test}

Table~\ref{tab:ablation} shows that the proposed method boosts the CIDEr scores on standard Transformer for image caption generation, indicating the proposed visual-semantic module and scattered-connection mechanism promote the CIDEr-D score in contrast with other baseline models. 

\begin{table}[thb]
\resizebox{\linewidth}{!}{%
\begin{tabular}{@{}lllllll@{}}
\toprule
Model & BLEU-1 & BLEU-4 & METEOR & ROUGE-L & CIDEr & SPICE \\ \midrule
Ours & 0.802 & 0.378 & 0.283 & 0.580 & \textbf{1.272} & 0.225 \\ 
w/o refinement & 0.786 & 0.366 & 0.277 & 0.570 & 1.202 & 0.210 \\
\bottomrule
\end{tabular}%
}
\caption{Comparison with standard Transformer.}
\label{tab:ablation}
\vskip -4mm
\end{table}

\begin{figure*}[thb]
\begin{center}
\includegraphics[width=\textwidth]{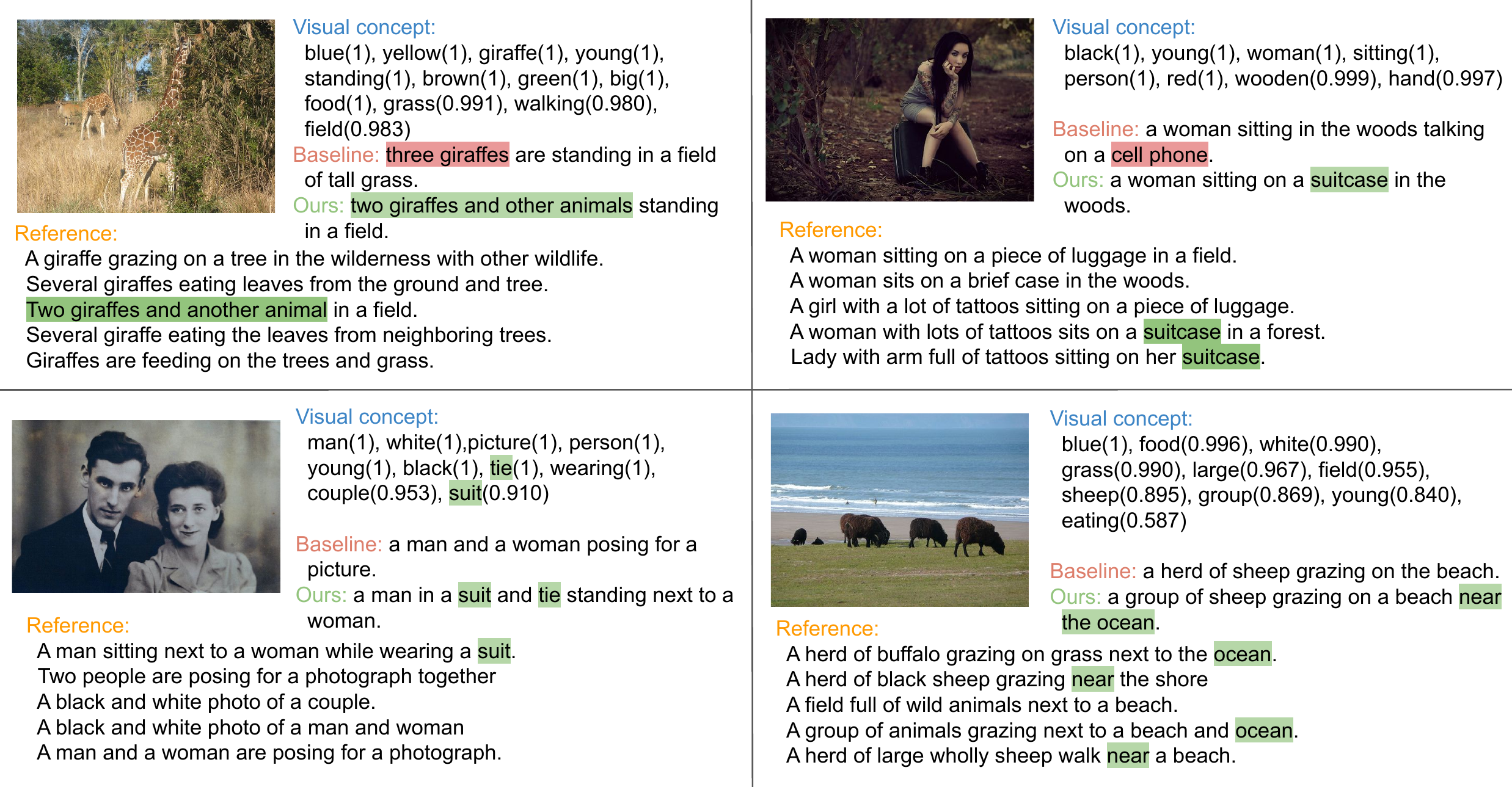}
\vskip -3mm
\caption{Detected tags and generated captions using baseline (Transformer) and proposed models on MS-COCO, where red and green backgrounds indicate wrong and correct predictions respectively. The value in brackets means the confidence (\emph{i.e.}, probability) of corresponding tags in the image.
}
\label{fig:case}
\end{center}
\vskip -5mm
\end{figure*}

\paragraph{Qualitative Analysis}
Fig.~\ref{fig:case} illustrates the generated caption samples, where the detected visual concepts advance the quality of predicted captions. As shown in Fig.~\ref{fig:case} bottom left, discovered visual tags enhance the adequacy of generated captions, such as ``tie'' with the confidence of 1, which is not mentioned in ground truth. Our model adds ``near the ocean'' as the adverbial modifier in comparison with baseline in the bottom right. 

{\bf Better Accuracy} Transformer baseline without the proposed method sometimes generates mismatched words, which can be reliably rectified by the proposed method. For example, in upper left of Fig.~\ref{fig:case}, our model correctly predicts the presence of ``two giraffes and another animal'' but baseline identifies them as ``three giraffes'' by mistake.

{\bf Better Adequacy} Our model can capture more specific details and meaningful contents in the image background that might be ignored by the baseline or even omitted in the ground truth. For example, as shown in Fig.~\ref{fig:case} (bottom left), the proposed model predicts the occurrence of ``tie'' which is overlooked by both the baseline and ground truth.

{\bf Implicit Visual Concept Modeling} We found that the detected concepts can exactly match semantic objects in the image. Notably, such visual-concept detection can be treated as the side effect of the scatter-connection mappings since the proposed method implicitly learns the visual mapping from the visual objects to visual-concept vocabulary (as aforementioned in Sec.~\ref{subsec:refine}), instead of using an explicit training objective for learning such mapping.

To investigate the necessity of explicit visual-concept learning, we design further experiments by pre-training the visual-concept detector for multi-label binary classification before training the caption generator. We empirically find it unnecessary to pretrain the visual-concept detector before the caption generation, as shown in Sec.~\ref{ap_subsec: ablation}.



\section{Conclusion}
In this work, we propose a new visual-concept based image captioning framework to generate meaningful descriptive sentences. Leveraging visual concepts and scatter mapping, \emph{RefineCap} demonstrates its effectiveness on the MS-COCO dataset. 


\bibliography{anthology,custom}
\bibliographystyle{acl_natbib}

\clearpage
\appendix
\section{Appendices}
\label{sec:appendix}

\subsection{Case Study of Generated Captions}
\label{ap_subsec: cases}
\begin{figure*}[htp]
  \centering
  \subfloat[]{\label{subfig:c2}\includegraphics[width=\textwidth]{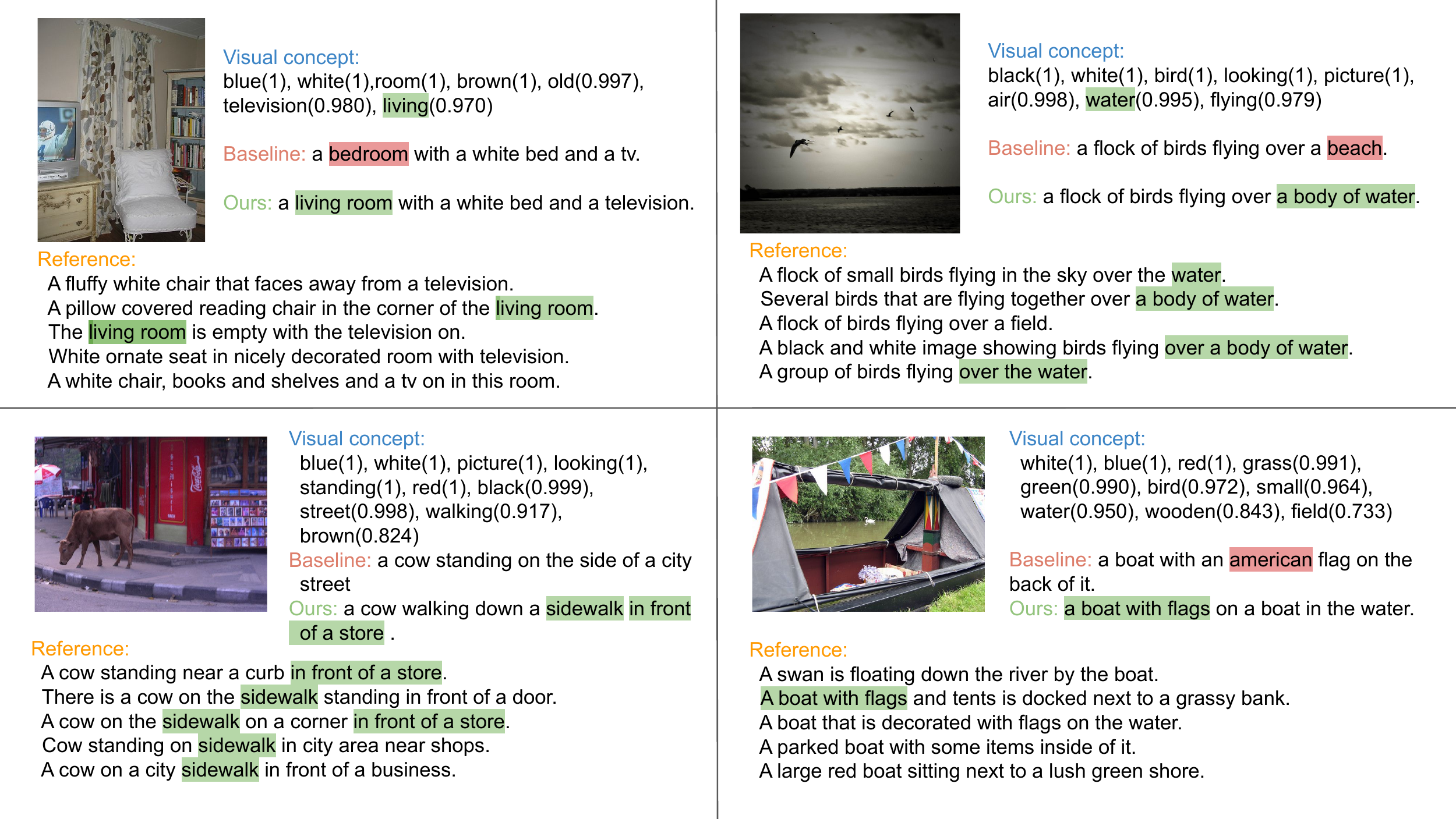}}
  \\
  \subfloat[]{\label{subfig:c3}\includegraphics[width=\textwidth]{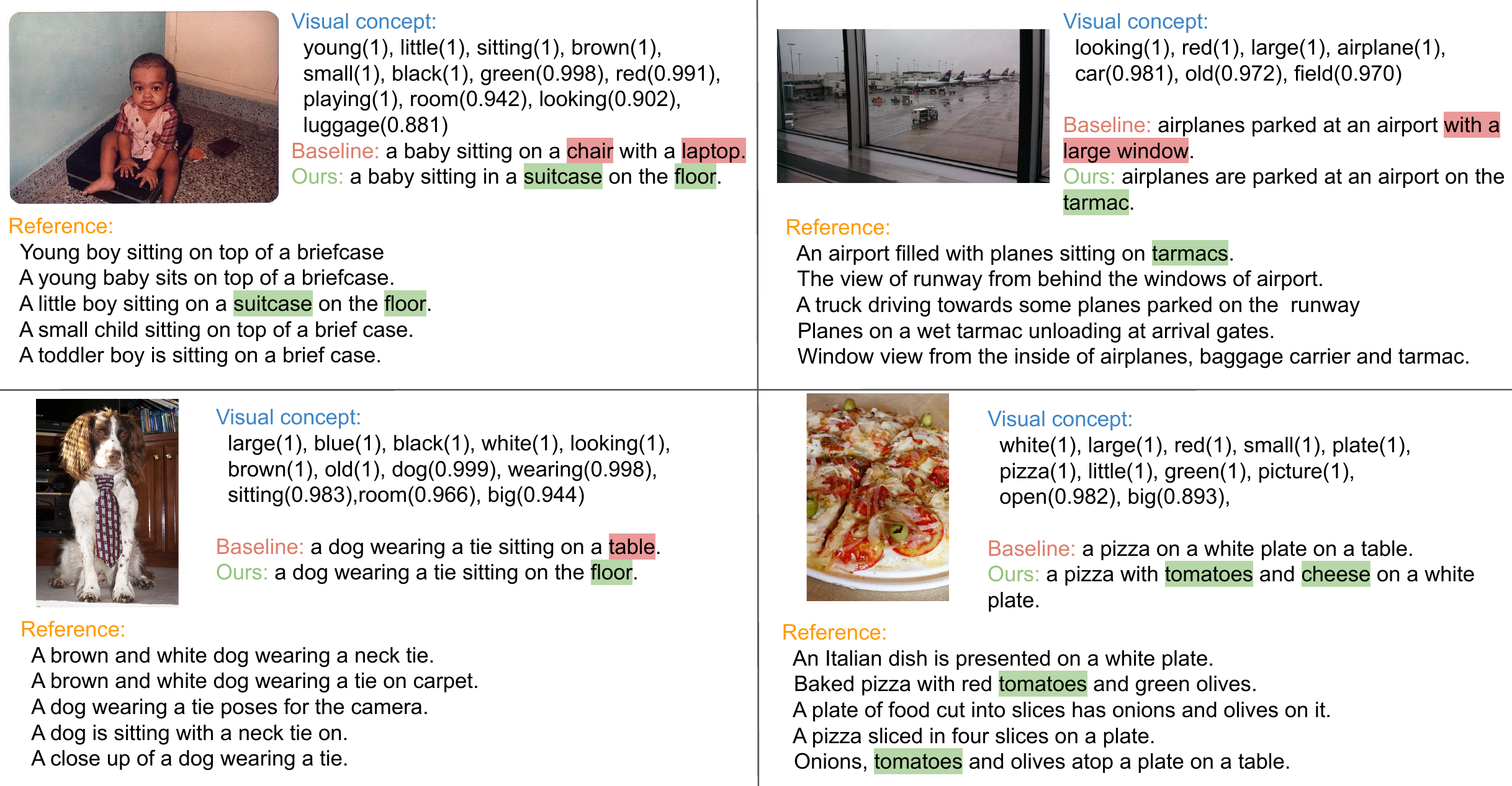}}
  \caption{Supplemental examples of the generated captions.}\label{fig:case2}
\end{figure*}

\begin{figure*}[htp]
  \centering
  \subfloat[]{\label{subfig:c4}\includegraphics[width=\textwidth]{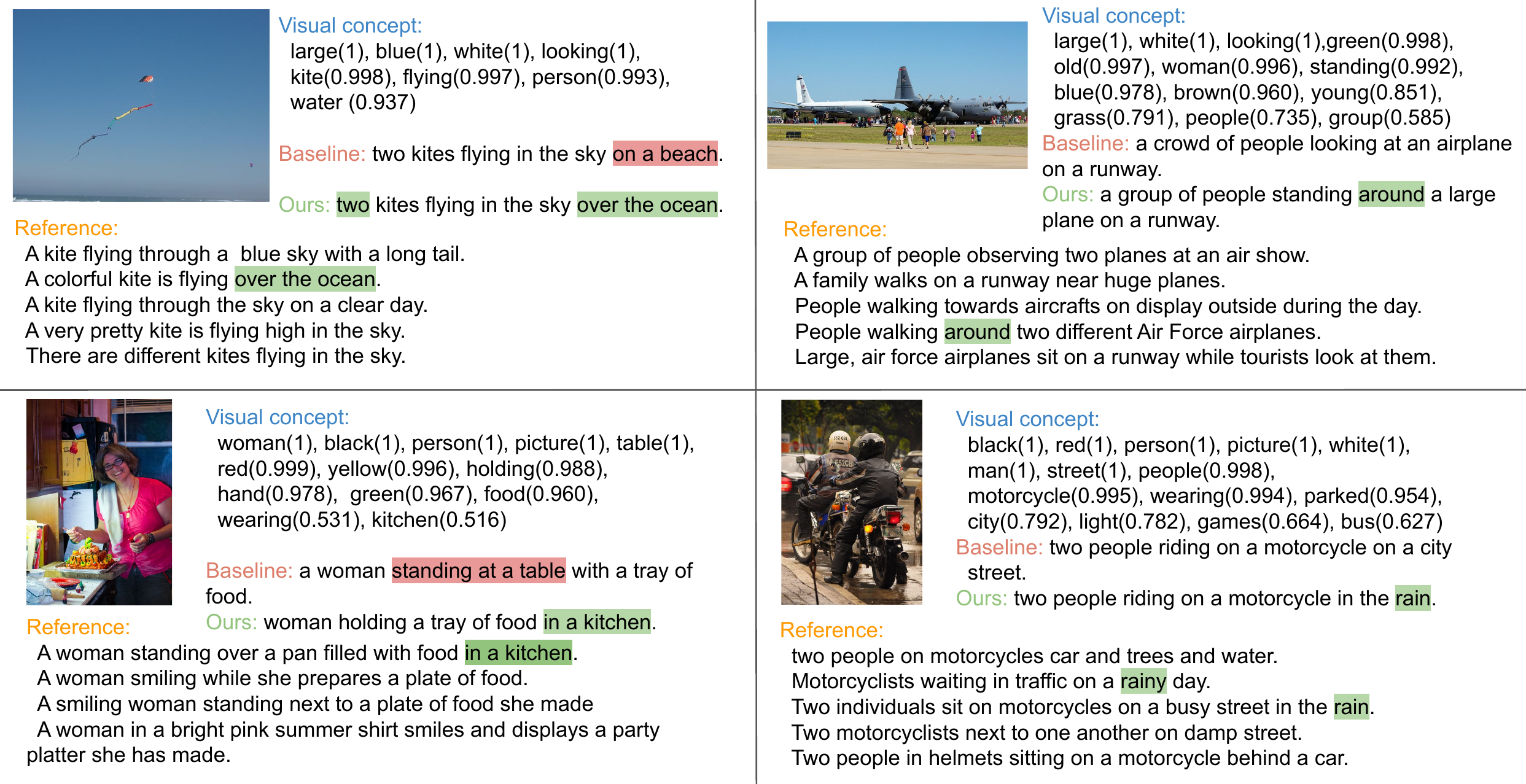}}
  \\
  \subfloat[]{\label{subfig:c5}\includegraphics[width=\textwidth]{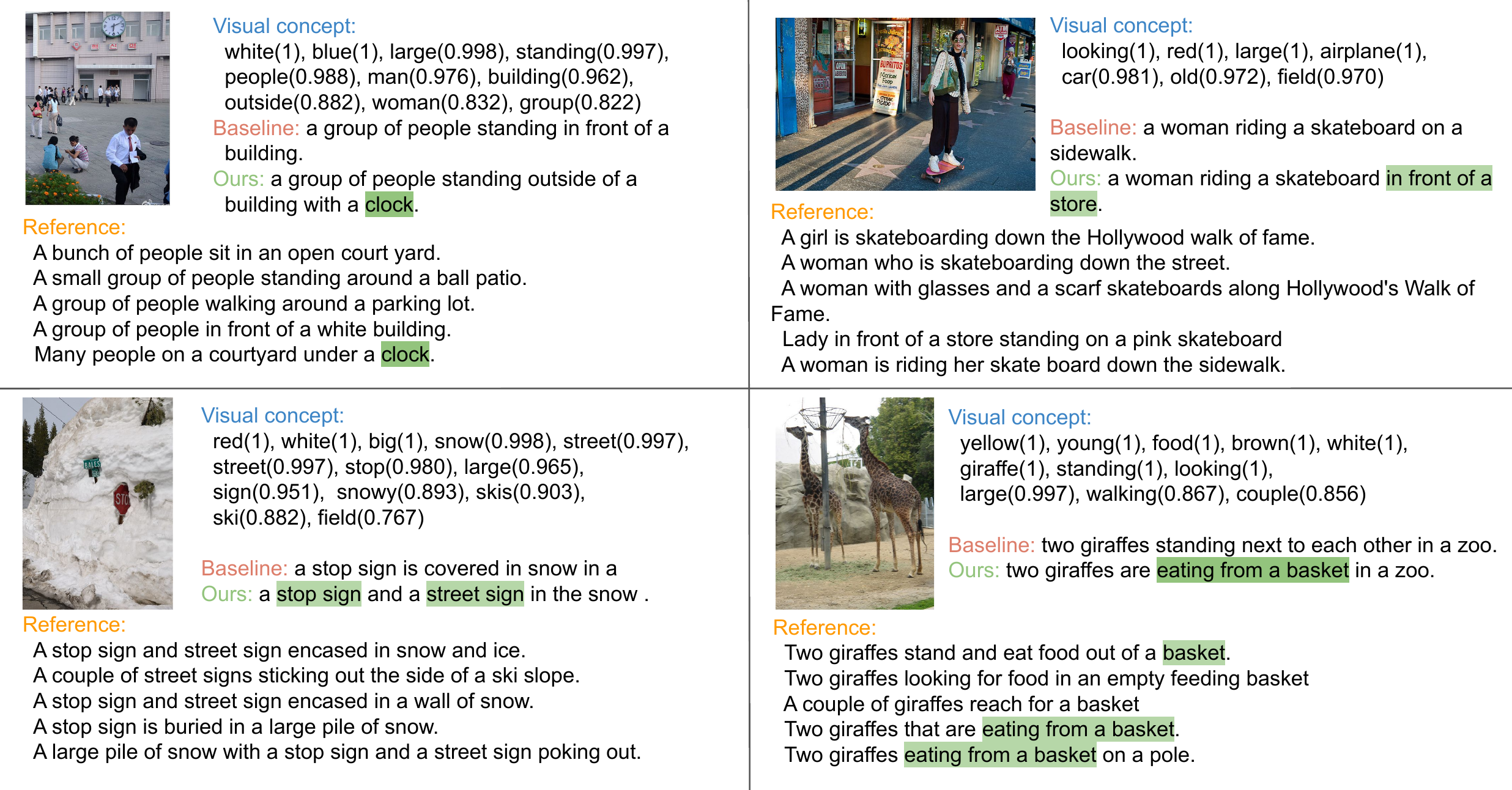}}
  \caption{Supplemental examples of the generated captions.}\label{fig:case3}
\end{figure*}

Fig.~\ref{fig:case},~\ref{fig:case2},~\ref{fig:case3} manifest the comparison between the baseline and proposed models, with the presence of visual concepts and their corresponding confidence (presented in brackets) for each image. 

\paragraph{Visual-Concept Indication} The proposed method implicitly learns the connection between visual concepts and image regions without explicit training on the visual concept detector. Further analysis can be found at Appendix~\ref{eq:tag_loss}.

\paragraph{Better Adequacy} It can be seen that our model can capture more specific details and meaningful contents in the image background that might be ignored by the baseline or even omitted in the ground truth. For example, as shown in the bottom left in Fig.~\ref{fig:case}, the proposed model predicts the occurrence of ``tie'' which is overlooked by both the baseline and ground truth. Other examples like ``ocean'' at the bottom right of Fig.~\ref{fig:case}, ``store'' at bottom left example of Fig.~\ref{subfig:c2}, ``ocean'' at upper left in Fig.~\ref{subfig:c4}, ``a ground of people ... around'' at the upper right~Fig.~\ref{subfig:c4}, ``kitchen'' at the bottom left of~Fig.~\ref{subfig:c4}, ``clock'' at upper left in Fig.~\ref{subfig:c5}, ``store'' at upper right in Fig.~\ref{subfig:c5}, ``street sign'' at bottom left in Fig.~\ref{subfig:c5}, ``basket'' at bottom right in Fig.~\ref{subfig:c5}, \emph{etc.} We found that captions generated by the proposed model could achieve better scores in terms of the adequacy.

\paragraph{Better Accuracy}
Meanwhile, captions decoded by the proposed models equip better accuracy. For instance, in upper left of Fig.~\ref{fig:case}, our model predicts the presence of ``two giraffes and another animal'' but baseline identifies them as ``three giraffes'' by mistake. Other proofs, such as ``suitcase'' in the top right of Fig.~\ref{fig:case}, ``living room'' in the top left of Fig.~\ref{subfig:c2}, `` a body of water'' in the top right of Fig.~\ref{subfig:c2}, ``flags'' in the bottom right of Fig.~\ref{subfig:c2}, ``suitcase'' and ``floor'' in the top left of Fig.~\ref{subfig:c3}, ``floor'' in the bottom left of Fig.~\ref{subfig:c3}, \emph{etc.}
Wrongly predicted words are marked in red in the examples. It can be concluded that the proposed model can produce image descriptions in a more accurate manner.


\subsection{Additional Details for Experiments}
We use SpaCy\footnote[3]{\url{https://spacy.io/}} as the tokenizer and pos tagger to process the reference captions, with lowercase and punctuation removal. Tokens whose occurrence less than 5 are treated as unknown words in our vocabulary. The reference vocabulary size is 10,201, whereas the most frequent 1,000 words from reference sentences are used as the visual-concept vocabulary. 
We set the maximum length of sentences as 20, in which all input sequences are right padded with 0s or cut on the right hand side. For hyperparameter tuning, we test the decoder layer number $N_\textrm{dec} \in \{3, 6\}$ and found that model with 6 layers slows down the training speed but achieves the similar performance.

RL training takes the majority of time costs due to the reward computation. Specifically, it took around 4/2/1.5 hours to train one epoch for RL/MLE/tag training in our experiments.

\subsection{Ablation Study Curves}
\label{ap_subsec: ablation}
\begin{figure*}[p]
\begin{center}
\includegraphics[width=\textwidth]{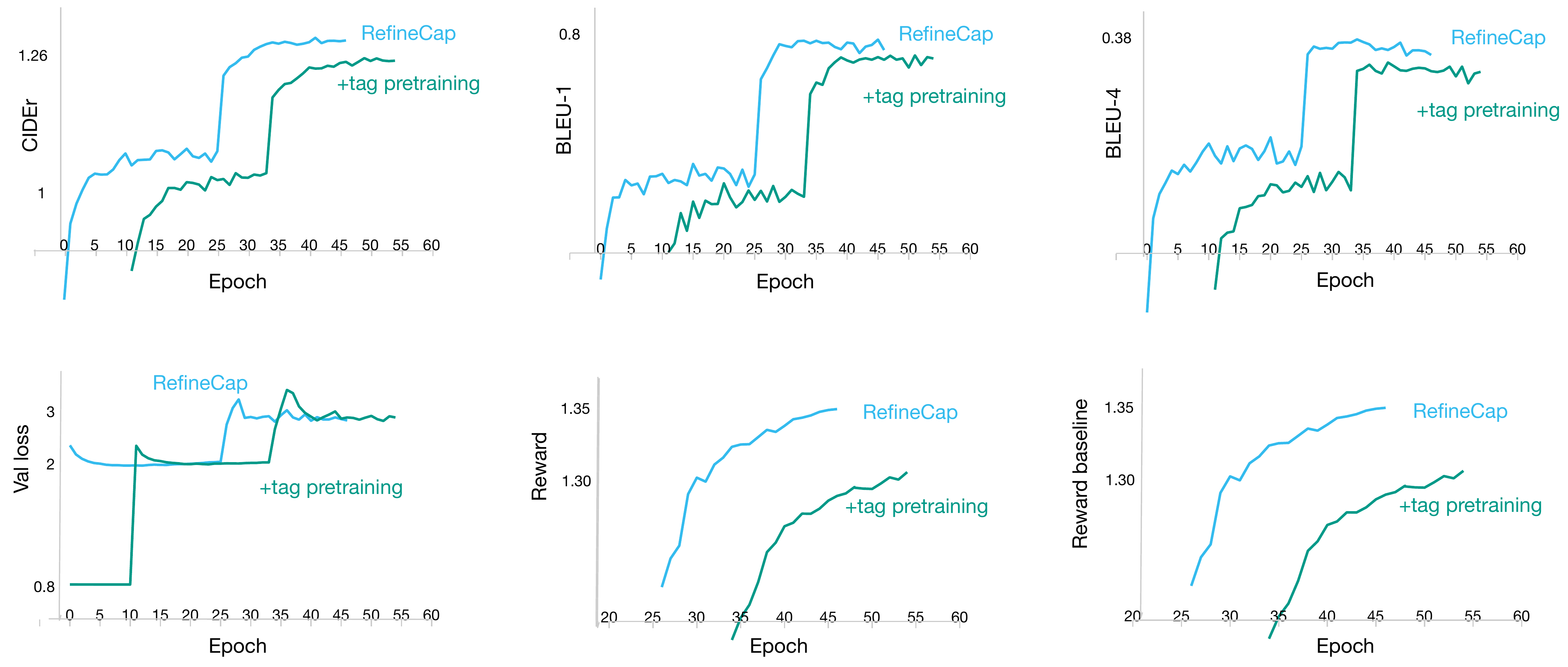}
\caption{The performance curves of \emph{RefineCap} w/ and w/o tag pretraining (in blue / green separately) on evaluation set, including CIDEr-D/BLEU-1/BLEU-4 metrics, validation loss, rewards and baseline reward values.}
\label{fig:curve}
\end{center}
\end{figure*}

To investigate the necessity of explicit training on visual tags, we add additional pretraining stage on the visual concept detection as a multi-label classification task.
Fig.\ref{fig:curve} compares the performance of proposed model with pretraining for 10 epochs using a binary cross-entropy loss (referred as \textit{+tag pretraining}), and the counterpart without the pretraining, \emph{i.e., RefineCap}, on the validation set. As shown in Fig.\ref{fig:curve}, \emph{RefineCap} conducts supervised training using cross-entropy loss as the weight initialization over the period of the beginning 25 epochs, followed by reinforcement learning process. Model with tag pretraining has three different stages: tag pretraining from epoch 0-10, supervised training from epoch 10-35, and finally reinforcement learning after 35-th epoch.

It can be seen that the overall trend of CIDEr-D scores, BLEU-1, BLEU-4 curves for models with tag pretraining witnesses the degeneration of performance, with no obvious change but a slight decrease, and the extra time and computing costs in the first 10 epochs. Besides, over the period of reinforcement learning, the rewards and reward baseline values of \textit{+tag pretraining} models have decreased with 0.05 shift due to the different starting points but maintain a similar increase slope as \emph{RefineCap}. Thus we extrapolate that explicit visual-concept pretraining could impede the weight initialization for the following training process in our model. We also observed a clear performance gain with REINFORCE algorithm in the final stage.


\end{document}